\pdfoutput=1

\documentclass[11pt]{article}

\usepackage[]{acl}

\usepackage{times}
\usepackage{latexsym}

\usepackage[T1]{fontenc}

\usepackage[utf8]{inputenc}

\usepackage{microtype}
\usepackage{amsmath}
\usepackage{amssymb}
\usepackage{graphicx}
\usepackage{bbm}
\usepackage{booktabs}
\usepackage{multirow}

\title{CSS: Combining Self-training and Self-supervised Learning for Few-shot Dialogue State Tracking}

\author{Haoning Zhang$^{1,3}$, Junwei Bao$^2$\thanks{~~Corresponding author}~, Haipeng Sun$^2$,  \\ \bf Huaishao Luo$^2$, Wenye Li$^4$, Shuguang Cui$^{3,1, 5}$\\
    $^1$FNii, CUHK-Shenzhen ~ $^2$JD AI Research\\
    $^3$SSE, CUHK-Shenzhen ~ $^4$SDS, CUHK-Shenzhen ~ $^5$Pengcheng Lab\\
    {\tt haoningzhang@link.cuhk.edu.cn, sunhaipeng6@jd.com,} \\
    {\tt \{baojunwei001, huaishaoluo\}@gmail.com,} \\ {\tt \{wyli, shuguangcui\}@cuhk.edu.cn}
}

\begin{document}
\maketitle
\begin{abstract}

Few-shot dialogue state tracking (DST) is a realistic problem that trains the DST model with limited labeled data. Existing few-shot methods mainly 
transfer knowledge learned from external labeled dialogue data (e.g., from question answering, dialogue summarization, machine reading comprehension tasks, etc.) into DST, 
whereas collecting a large amount of external labeled data is laborious,
and the external data may not effectively contribute to the DST-specific task.
In this paper, we propose a few-shot DST framework called $\textbf{CSS}$, which $\textbf{C}$ombines $\textbf{S}$elf-training and $\textbf{S}$elf-supervised learning methods. The unlabeled data of the DST task is incorporated into the self-training iterations, where the pseudo labels are predicted by a DST model trained on limited labeled data in advance. Besides, a contrastive self-supervised method is used to learn better representations, where the data is augmented by the dropout operation to train the model. Experimental results on the MultiWOZ dataset show that our proposed CSS achieves competitive performance in several few-shot scenarios.

\end{abstract}

\section{Introduction}

Dialogue state tracking (DST) is an essential sub-task in a task-oriented dialogue system \citep{yangubar, ramachandran2022caspi, sun-etal-2022-bort}. It predicts the dialogue state corresponding to the user's intents at each dialogue turn, which will be used to extract the preference and generate the natural language response \citep{williams2007partially, young2010hidden, lee2016dialog, mrkvsic2017neural, xu2018end, wu2019transferable, kim2020efficient, ye2021slot, wang-etal-2022-luna}.
Figure \ref{diaexp} gives an example of DST in a conversation, where the dialogue state is accumulated and updated after each turn. 

\begin{table}[!t]
\centering
\small
\scalebox{0.95}{\begin{tabular}{@{}ll@{}}
\toprule
\multicolumn{2}{l}{\begin{tabular}[c]{@{}l@{}}Usr: Hi I am looking for a \textbf{restaurant in the north} that \\ \qquad serves \textbf{Asian} oriental food.\\ Sys: I would recommend \textbf{Saigon city}. Would you like to \\ \qquad make a reservation?\\ Usr: That sounds great! We would like a reservation for \\ \qquad \textbf{Monday} at \textbf{16:45} for \textbf{6 people}. Can I get the reference \\ \qquad number for our reservation?\end{tabular}} \\ \midrule
\multicolumn{2}{l}{\begin{tabular}[c]{@{}ll@{}}restaurant-area-north &  restaurant-food-Asian \\ restaurant-name-Saigon city & restaurant-book day-Monday \\ restaurant-book time-16:45 & restaurant-book people-6\end{tabular}} \\ \bottomrule
\end{tabular}}
\caption{A dialogue example containing utterances from user and system sides and the corresponding dialogue state (a set of domain-slot-value pairs).}
\label{diaexp}
\end{table}

Training a DST model requires plenty of dialogue corpus containing dialogue utterances and human-annotated state labels, whereas annotating is costly. Therefore, the DST models are expected to have acceptable performance when trained with limited labeled data, i.e., in the few-shot cases \citep{wu2020improving}. Previous studies on few-shot DST solve the data scarcity issue mainly by leveraging external labeled dialogue corpus to pre-train the language models, which are then transferred into the DST task \citep{wu2020tod, su-etal-2022-multi, shin2022dialogue}. However, there exist several disadvantages: first, collecting a large amount of external labeled data is still laborious; second, utilizing the external data is heavily dependent on computational resources since the language models have to be further pre-trained; third, the external data always comes from different conversation scenarios and NLP tasks, such as dialogues in multi topics, question answering, dialogue summary, etc. The data types and distributions differ from the DST-specific training data, making it less efficient to transfer the learned knowledge into DST.

\begin{figure*}[!t]
	\centering
	\includegraphics[width=14cm]{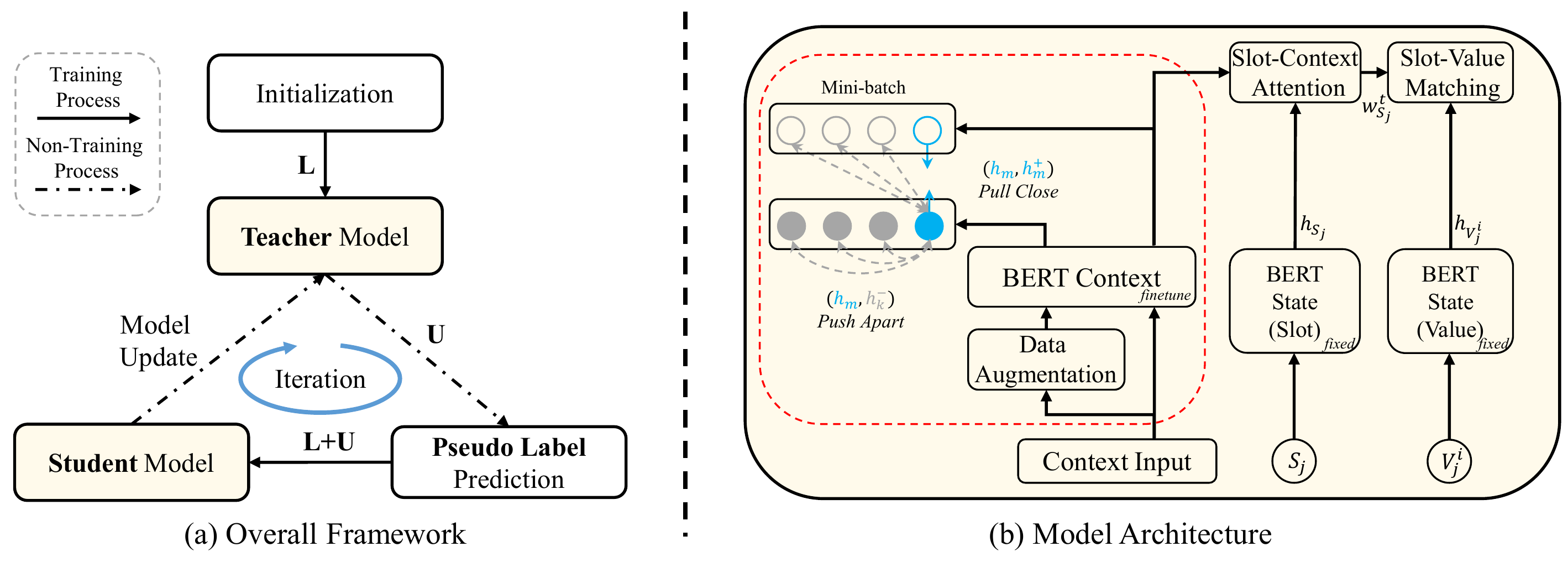}
	\caption{\textcolor{black}{The description of CSS. Part (a) is the overall teacher-student training iteration process, \textbf{L} and \textbf{U} correspond to labeled and unlabeled data. Part (b) is the model architecture for both teacher and student, where the red dashed box is the illustration of the self-supervised learning object through the dropout augmentation: narrow the distance between each instance and its corresponding augmented one (pull close), enlarge its distance to the rest in the same batch in representation area (push apart).}}
	\label{STframe}
\end{figure*}

We consider utilizing the unlabeled data of the DST task, which is easy to access and has similar contents to the limited labeled data, so that the DST model can be enhanced by training on an enlarged amount of data corpus. In this paper, we propose a few-shot DST framework called CSS, which \textbf{C}ombines the \textbf{S}elf-training and \textbf{S}elf-supervised methods. Specifically, a DST model is first trained on limited labeled data and used to generate the pseudo labels of the unlabeled data; then both the labeled and unlabeled data can be used to train the model iteratively. Besides, we augment the data through the contrastive self-supervised dropout operation to learn better representations. Each training instance is masked through a dropout embedding layer, which will act as the contrastive pair, and the model is trained to pull the original and dropout instances closer in the representation area. Experiments on the multi-domain dialogue dataset MultiWOZ demonstrate that our CSS achieves competitive performance with existing few-shot DST models.

\section{Related Work}
Few-shot DST focuses on the model performance with limited labeled training data, which overcomes the general data scarcity issue.
Existing DST models enhance the few-shot performance mainly by incorporating external data of different tasks to further pre-train a language model, which is still collection and computational resources demanding \citep{gao2020machine, lin2021zero, su-etal-2022-multi, shin2022dialogue}. Inspired by self-training that incorporates predicted pseudo labels of the unlabeled data to enlarge the training corpus \citep{wang2020combining, mi2021self, sun2021self}, in this paper, we build our framework upon the NoisyStudent method \citep{xie2020self} to enhance the DST model in few-shot cases.

Self-supervised learning trains a model on an auxiliary task with the automatically obtained ground-truth \citep{mikolov2013efficient, jin2018explicit, wu2019self, devlin2019bert, lewis2020bart}. As one of the self-supervised approaches, contrastive learning succeeds in various NLP-related tasks, which helps the model learn high-quality representations \citep{cai2020group, klein2020contrastive, gao2021simcse, yan-etal-2021-consert}. In this paper, we construct contrastive data pairs by the dropout operation to train the DST model, which does not need extra supervision.

\section{Methology}
Figure \ref{STframe} shows the CSS framework, where (a) is the overall training framework, and (b) is the architecture of both teacher and student models. Our CSS follows the NoisyStudent self-training framework \citep{xie2020self}. After deriving a teacher DST model trained with labeled data, it's continuously trained and updated into the student DST model with both labeled and unlabeled data, where the pseudo labels of the unlabeled data are synchronously predicted. Unlike the original NoisyStudent augmenting training data only in the student training stage, we implement the contrastive self-supervised learning method in both training teacher and student models, \textcolor{
black}{where each training instance is augmented through a dropout operation, and the model is trained to group each instance with its augmented pair closer and diverse it far
from the rest in the same batch.}

\subsection{DST Task and Base Model}

Let's define $D_t = \{(Q_t, R_t)\}_{t=1:T}$ as the set of system query and user response pairs in total $T$ turns, $B_t$ as the dialogue state for each dialogue turn, which contains a set of (domain-slot $S$, value $V$) pairs: $B_t=\{(S_j,V^i_j)|1 \leq j \leq J, 1 \leq i \leq I \}$, assuming there are $J$ (domain-slot) pairs, and $\mathcal V_j=\{V^i_j\}$ is the value space of slot $S_j$ with $I$ candidates. DST task aims to generate the dialogue state at the $t$-th turn $B_t$, given all the dialogue utterances and the predicted state from the previous turn.

The base DST model is a standard BERT-based matching framework training on a small dataset \citep{ye2022assist}, denoted as \textbf{BASE}. The context input is the concatenation of the dialogue utterances and state from the previous turn, denoted as $C_t = [CLS] \oplus D_1 \oplus ... \oplus D_{t-1} \oplus B_{t-1} \oplus [SEP] \oplus D_t \oplus [SEP]$; a BERT context encoder encodes the context input, denoted as $H_{t} = BERT_{finetune}(C_t)$
; for slots and values, another BERT state encoder with fixed parameters is used to derive the representations: $h_{S_j}=BERT_{fixed}(S_j),\quad h_{V_j^i}=BERT_{fixed}(V^i_j).$ During training, the parameters of the BERT state encoder will not be fine-tuned. For each slot, its context-relevant feature is derived through the multi-head attention, where the slot representation acts as query, the context representation acts as both the key and value \citep{vaswani2017attention}: $r^t_{S_j}=MultiHead(h_{S_j}, H_t, H_t)$. Then it's transformed by a linear and normalization layer: $w^{t}_{S_j} = LayerNorm(Linear(r^t_{S_j}))$, which is used to calculate the distance with each value representation of $S_j$, and the one with the smallest distance will be selected. The probability of selecting the ground truth $h_{V_j^{i'}}$ is denoted as:
\begin{equation}
	\resizebox{0.95\hsize}{!}{
		$P(V^{i'}_j|C_t, S_j) = \frac{exp(-|| w^{t}_{S_j}-h_{V_j^{i'}} ||_2)}{\sum_{V_j^i \in \mathcal V_j}exp(-||w^t_{S_j}-h_{V_j^i}||_2)}$, }
		\label{prob}
\end{equation}
and the DST objective is to minimize the sum of the negative log-likelihood among the $J$ slots:
\begin{equation} 
	\textcolor{black}{L_{d} = \sum_{j=1}^J -log(P(V^{i'}_j|C_t, S_j))}.
	\label{baseloss}
\end{equation} 
We implement our CSS built on the model BASE, and it's also available to transfer CSS into other DST-related models.

\subsection{Self-training}
Let $L$, $U$ be labeled and unlabeled data,  $\textbf{X}=\{x_n\}_{n=1:N}$ be the set of training instances containing $N$ dialogues. A teacher $f_T$ is trained with $L$; then for each $x_n \in U$, the dialogue state is predicted by $f_T$ and acts as the pseudo label.
Both $L$ (with ground labels) and $U$ (with pseudo labels) are used to train a student $f_S$ with the following objective function:

\begin{equation} 
\begin{aligned}
	\textcolor{black}{L^{*}_{d} = \sum_{j=1}^J -log(P(V^{i^*}_j|C_t, S_j))} \\
	\textcolor{black}{+\sum_{j=1}^J -log(P(V^{i'}_j|C_t, S_j))}
\end{aligned}
\end{equation} 
$V^{i^*}_j$ and $V^{i'}_j$ correspond to the pseudo and ground labels from $U$ and $L$. $f_S$ will replace $f_T$ to re-predict the pseudo labels on $U$, and the training-prediction-training loop will iterate until $f_S$ converges.

\subsection{Self-supervised Learning}
We implement the contrastive self-supervised method to learn better representations, where a simple yet effective dropout operation augments the training instances.
Denote $\{x_m\}_{m=1:M}$ as the training instances in a batch with size $M$. Each $x_m$ is augmented into $x_m^+$ through a dropout embedding layer, and both of them are encoded by the BERT context encoder: $h_m = BERT_{finetune}(x_m), \quad h_m^+ = BERT_{finetune}(x_m^+)$. Then the model is trained to narrow their representation distances with the contrastive objective:
\begin{equation}
    \resizebox{0.95\hsize}{!}{
    \textcolor{black}{$L_m = -log \frac{e^{sim(h_m, h_m^+)/\tau}}{e^{sim(h_m, h_m^+)/\tau}+\sum\limits_{k=1}^{2M-2}(e^{sim(h_m, h_k^-)/\tau})}$}},
\end{equation}
where $\tau$ is the temperature parameter, and $\{h_k^-\}$ correspond to all training instances in the same batch except $h_m$ and $h_m^+$ ($2M-2$ instances). In simpler words, each training instance and its dropout pair are treated as the ones having similar semantic representations. 

\begin{table*}[t]
\centering
\scalebox{0.80}{\begin{tabular}{@{}l|c|cccc|c@{}}
\toprule
Models & Pre-trained Model ($\#$ Params.) & 1\% & 5\% & 10\% & 25\% & 100\% \\ \midrule
TRADE \citep{wu2019transferable} & - & 9.70  & 29.38  & 34.07 & 41.41 & 48.62\\
MinTL \citep{lin2020mintl} & BART-large (400M)  & 9.25 & 21.28 & 30.32 & - & 52.10\\
TRADE$_{ssup}$ \citep{wu2020improving}& - & 20.41  & 33.67  & 37.16  & 42.69 & 48.72 \\
STAR \citep{ye2021slot}& BERT-base (110M) & 8.08 & 26.41 & 38.45 & 48.29 & 54.53\\
TOD-BERT$^*$ \citep{wu2020tod} & BERT-base (110M) & 10.30 & 27.80 & 38.80 & 44.30 & - \\
PPTOD$^*$ \citep{su-etal-2022-multi} & T5-large (770M)  & 31.46 & 43.61 & 45.96 & 49.27 & 53.89\\
DS2$^*$ \citep{shin2022dialogue} & T5-large (770M)  & \textbf{36.15} & \textbf{45.14} & 47.61 & 50.45 & 54.78 \\ \midrule
BASE & BERT-base (110M) & 13.19 & 37.19 & 44.23 & 49.20 & 53.97 \\
\textbf{CSS} & BERT-base (110M) & 14.06 & 41.90 & \textbf{47.96} & \textbf{51.88}  & \textbf{55.02} \\ \bottomrule
\end{tabular}}
\caption{Joint goal accuracy on MultiWOZ 2.0. $^*$ means the model incorporates external labeled dialogue data to pre-train a language model. The results of TRADE and TOD-BERT come from \citet{wu2020improving}; MinTL comes from \citet{su-etal-2022-multi}; STAR, 25\% of PPTOD and DS2 are reproduced by using their released codes.}
\label{jga}
\end{table*}

\subsection{\textcolor{black}{Optimization}}
\textcolor{black}{Besides the CSS and BASE models, another two ablations on BASE are conducted: BASE w/ SSL and BASE w/ ST. We first train a BASE model:
$L_{BASE}=L_{d}$, then we train a BASE model adding the self-supervised method, denoted as BASE w/ SSL: $L_{SSL}=L_{d} + L_m$.
Next the unlabeled data is incorporated, and we train a BASE model adding the self-training iterations, denoted as BASE w/ ST: $L_{ST}=L^{*}_{d}$, and finally we train the CSS model: $L_{CSS} = L^{*}_{d} + L_m$.
The performance of the four models will be shown in Section 4.}

\section{Experiments}
\textcolor{black}{In this section, we first give the experimental dataset and training details, then show the experimental results compared with several existing baselines, ablation studies in both multi-domain and single-domain accuracy, and the error analysis.}
\subsection{Dataset and Few-shot Settings}
We evaluate our CSS on MultiWOZ 2.0 \citep{budzianowski2018multiwoz}, a task-oriented dialogue dataset containing 7 domains (attraction, hospital, hotel, police, restaurant, taxi, train) and around 8400 multi-turn training dialogues. Since the hospital and police domains do not have dialogues in validation and test sets, we follow the previous work \citep{wu2019transferable} to use five domains (attraction, hotel, restaurant, taxi, train) as training data with 30 (domain, slot) pairs. We randomly select $1\%$, $5\%$, $10\%$, and $25\%$ labeled training data to simulate the few-shot cases. For self-training, the amount of unlabeled data is 50\% of the training dataset in MultiWOZ 2.0 and excluded from the labeled training data. For each case, we use three different fixed random seeds during the whole data selection and training process, and the final result is averaged. We use the joint goal accuracy to evaluate the model, which is the ratio of dialogue turns that all the (domain-slot-value) pairs are correctly predicted. 

\subsection{Training Details}
We choose BERT-base-uncased as the context encoder.
The batch size is set to $8$. The AdamW optimizer is applied to optimize the model with the learning rate 4e-5 and 1e-4 for encoder and decoder \citep{loshchilov2018decoupled}.
Both the dropout rate and the temperature parameter are set to $0.1$. All the models are trained on a single P40. For the sake of the computation resources 
efficiency, each teacher DST model is trained for 50 epochs, and each student model is trained over 3 iteration loops with 10 epochs for each loop.
\label{traindetail}

\subsection{Main Results}
Table \ref{jga} shows the results in terms of joint goal accuracy. Our CSS generally performs well in the four few-shot settings, especially achieving SOTA results using 10\% and 25\% training data. Besides, CSS outperforms all the methods using 100\% labeled training data, where all the labeled dialogues are used to train a teacher model, and the student model is trained on 150\% data, 50\% of which has both labels and pseudo labels. It's also observed that when using 1\% and 5\% training data, PPTOD and DS2 perform better than others. Specifically, both PPTOD and DS2 use the T5-large language model \citep{raffel2020exploring}, which has a remarkable contribution to the prediction accuracy, especially when the amount of labeled DST data is strictly limited. Besides, PPTOD pre-trains T5 on various dialogue-related tasks and data, and DS2 also pre-trains T5 on dialogue summarization data, which further
enhance their DST models by the dialogue-related knowledge. Therefore, compared with them, we conclude that the superiority of our CSS mainly comes from efficiently utilizing the DST-related unlabeled data, instead of the large language model or external dialogue data.

\begin{table}[t]
\centering
\scalebox{0.80}{\begin{tabular}{@{}l|cccc@{}}
    \toprule
     & 1\% & 5\% & 10\% & 25\% \\ \midrule
    BASE & 13.19 & 37.19 & 44.23 & 49.20 \\
    \quad w/ SSL & 13.26 & 38.73 & 44.87 & 49.48 \\
    \quad w/ ST & 14.07 & 40.33 & 47.00 & 51.78 \\ \midrule
    CSS & 14.06 & 41.90 & 47.96 & 51.88 \\ \bottomrule
\end{tabular}}
\caption{Joint goal accuracy on MultiWOZ 2.0 of CSS and three ablations: BASE, BASE w/ SSL (self-supervised learning), BASE w/ ST (self-training).}
\label{ablation}
\end{table}

\begin{table}[t]
\centering
\scalebox{0.75}{
\begin{tabular}{@{}l|ccccc@{}}
    \toprule
    5\% & attraction & hotel & restaurant & taxi & train \\ \midrule
    BASE & 60.26 & 48.80 & 53.17 & 63.12 & 72.60\\
    \quad w/ SSL & 60.98 & 50.74 & 53.22 & 62.88 & 75.24\\
    \quad w/ ST & 61.34 & 51.96 & 55.69 & 63.12 & 77.41 \\ \midrule
    CSS & 62.97 & 53.27 & 57.32 & 63.55 & 79.13 \\ \bottomrule
\end{tabular}}
\caption{Domain joint accuracy using 5\% labeled training data.}
\label{5data}
\end{table}

\begin{table}[t]
\small
\centering
\begin{tabular}{@{}l||cc|cc@{}}
    \toprule
    Error Type & Ground & Prediction & Count & Ratio \\ \midrule
    \uppercase\expandafter{\romannumeral1} & \textit{active} & \textit{none} & 88 & 43.35\%  \\ 
    \uppercase\expandafter{\romannumeral2} & \textit{none} & \textit{active} & 68 & 33.50\% \\ 
    \uppercase\expandafter{\romannumeral3} & \textit{active} & \textit{active} & 47 & 23.15\%  \\ \bottomrule
\end{tabular}
\caption{The comparison of three wrong types and the number and corresponding ratio of wrong predictions in 100 sampled turns (203 wrong predictions in total).}
\label{error_count}
\end{table}

\begin{table}[t]
\small
\centering
\scalebox{0.85}{\begin{tabular}{@{}l|l@{}}
    \toprule
    Usr (turn 1) & I need a taxi at \textbf{Lan Hong House} to leave by \textbf{14:45}. \\ \midrule
    Sys (turn 2) & Okay, what is your destination?  \\ 
    Usr (turn 2) & I want to go to the \textbf{Leicester} train station. \\ \midrule 
    Sys (turn 3) & Have you in a white Honda, 07040297067 is the \\ 
        & phone number. \\
    Usr (turn 3) & Thanks for the quick response.    \\ \midrule \midrule
    Ground & \textcolor{blue}{taxi-departure-Lan Hong House} \\
           & \textcolor{blue}{taxi-leaveat-14:45}\\
           & \textcolor{red}{taxi-destination-Leicester} \\  \midrule
    Prediction & \textcolor{orange}{train-destination-Leicester}, \\
               & \textcolor{red}{taxi-destination-Autumn House} \\ \bottomrule
\end{tabular}}
\caption{A dialogue example containing three turns (divided by short lines) and the wrong predicted dialogue state at the third turn (the slots with value \textit{none} are omitted). The \textcolor{blue}{blue}, \textcolor{orange}{orange}, \textcolor{red}{red} domain-slot-value pairs correspond to the wrong type \uppercase\expandafter{\romannumeral1}, \uppercase\expandafter{\romannumeral2}, \uppercase\expandafter{\romannumeral3}. }
\label{errexp}
\end{table}

\subsection{Ablation Studies}
Table \ref{ablation} shows the performance of four models: BASE, BASE w/ SSL, BASE w/ ST, and CSS, in terms of joint goal accuracy. Table \ref{5data} shows the joint accuracy for every single domain using 5\% training data. It can be observed that in both two tables, BASE w/ SSL and BASE w/ ST perform better than BASE, and CSS gets the best accuracy, indicating the effectiveness of each individual method. Detailed experiment results are shown in \ref{wholeresult}.

\subsection{Error Analysis}
\textcolor{black}{We further analyze the wrong prediction types. There are 3 wrong types. Type \uppercase\expandafter{\romannumeral1} means the model fails to predict a correct (domain-slot-value) pair (the predicted value is \textit{none} while the ground truth is not, denoted as \textit{active}), Type \uppercase\expandafter{\romannumeral2} means the model predicts a value not contained in the ground truth (the ground truth value is \textit{none}), and Type \uppercase\expandafter{\romannumeral3} means the model predicts a value different from the ground truth (both the predicted and ground truth value are \textit{active}). We use the CSS model trained on 5\% labeled training data to make predictions on the testset. Among the dialogue turns containing wrong predictions, we randomly sample 100 turns and then sum all the wrong predicted domain-slot-value pairs, which is 203 in total. Table \ref{error_count} shows the comparison of three wrong types, the number of each wrong type pairs and the corresponding ratio, where Type \uppercase\expandafter{\romannumeral1} is the most common case. Table \ref{errexp} gives a three-turn dialogue example containing wrong predictions. This indicates that the prediction performance can be further enhanced by better modeling the dialogue context from history turns, and we leave it in further studies.}

\section{Conclusion}
In this paper, we propose CSS, a training framework combining self-training and self-supervised learning for the few-shot DST task. The self-training enlarges the training data corpus by incorporating unlabeled data with pseudo labels to train a better DST model, and the contrastive self-supervised learning method helps learn better representations without extra supervision. Compared with the previous methods leveraging knowledge learned from a large amount of external labeled dialogue data, CSS is superior in smaller data scales and less computational resources. Experiments on MultiWOZ 2.0 demonstrate the effectiveness of CSS in several few-shot scenarios.

\section*{Acknowledgements}
The work was supported in part by the Basic Research Project No. HZQB-KCZYZ-2021067 of Hetao Shenzhen-HK S\&T Cooperation Zone, the National Key R\&D Program of China with grant No. 2018YFB1800800, by the National Key Research and Development Program of China under Grant No. 2020AAA0108600, by Shenzhen Outstanding Talents Training Fund 202002, by Guangdong Research Projects No. 2017ZT07X152 and No. 2019CX01X104, and by the Guangdong Provincial Key Laboratory of Future Networks of Intelligence (Grant No. 2022B1212010001).

\section*{Ethics}
The experiment dataset is publicly available and does not contain any sensitive information. We declare that all authors of this paper acknowledge the ACM Code of Ethics.

\bibliography{custom_new}
\bibliographystyle{acl_natbib}
\appendix

\section{Appendices}

\subsection{Experiment Results}
Tables \ref{BASE}, \ref{sup}, \ref{st}, \ref{s2} show all the experiments in joint goal accuracy of the four models: BASE, BASE w/ SSL, BASE w/ ST, CSS. Each of them is run on three random seeds for the four few-shot data ratio settings, and the final accuracy is averaged. Tables \ref{15avg}, \ref{1025avg} show the domain joint accuracy, and Tables \ref{1shot}, \ref{5shot}, \ref{10shot}, \ref{25shot} show detailed experiments in domain joint accuracy on three different seeds. 

\label{wholeresult}

\newpage

\begin{table}[!t]
\centering
\scalebox{0.95}{\begin{tabular}{@{}l|ccc|c@{}}
    \toprule
    Ratio & 1-run & 2-run & 3-run & Average \\ \midrule
    1\% & 13.17 & 13.42 & 12.99 & 13.19 \\
    5\% & 35.82 & 37.73 & 38.02 & 37.19 \\
    10\% & 43.43 & 44.00 & 45.26 & 44.23 \\
    25\% & 50.46 & 48.19 & 48.95 & 49.20 \\ \bottomrule
\end{tabular}}
\caption{Joint goal accuracy of BASE.}
\label{BASE}
\end{table}

\begin{table}[!t]
\centering
\scalebox{0.95}{\begin{tabular}{@{}l|ccc|c@{}}
    \toprule
    Ratio & 1-run & 2-run & 3-run & Average\\ \midrule
    1\% & 12.08 & 13.40 & 14.29 & 13.26 \\
    5\% & 36.50 & 39.71 & 39.97 & 38.73 \\
    10\% & 44.43 & 45.39 & 44.80 & 44.87 \\
    25\% & 49.12 & 50.12 & 49.21 & 49.48  \\ \bottomrule
\end{tabular}}
\caption{Joint goal accuracy of BASE w/ SSL.}
\label{sup}
\end{table}

\begin{table}[!t]
\centering
\scalebox{0.95}{\begin{tabular}{@{}l|ccc|c@{}}
    \toprule
    Ratio & 1-run & 2-run & 3-run & Average \\ \midrule
    1\% &13.80 & 14.60 &13.82 &  14.07\\
    5\% & 40.74 & 40.23 & 40.01 & 40.33 \\
    10\% & 47.26 & 46.92 & 46.82 & 47.00 \\
    25\% & 51.87 & 51.66 & 51.81 & 51.78 \\ \bottomrule
\end{tabular}}
\caption{Joint goal accuracy of BASE w/ ST.}
\label{st}
\end{table}

\begin{table}[!t]
\centering
\scalebox{0.95}{\begin{tabular}{@{}l|ccc|c@{}}
    \toprule
    Ratio & 1-run & 2-run & 3-run & Average \\ \midrule
    1\% & 12.35  & 14.52  & 15.32 & 14.06\\
    5\% & 40.93  & 42.24 & 43.09 & 41.90 \\
    10\% & 47.87 & 48.63 & 47.39 & 47.96 \\
    25\% & 51.59 & 52.43 & 51.64 & 51.88 \\ \bottomrule
\end{tabular}}
\caption{Joint goal accuracy of CSS.}
\label{s2}
\end{table}

\begin{table*}[!t]
\centering
\scalebox{0.785}{
\begin{tabular}{@{}l|ccccc||l|ccccc@{}}
    \toprule
    1\% & attraction & hotel & restaurant & taxi & train & 5\% & attraction & hotel & restaurant & taxi & train \\ \midrule
    BASE & 43.38 & 32.49 & 33.11 & 58.47 & 25.67 & BASE & 60.26 & 48.80 & 53.17 & 63.12 & 72.60 \\ 
    \quad w/ SSL & 43.35 & 32.60 & 31.34 & 58.49 & 25.70 & \quad w/ SSL & 60.98 & 50.74 & 53.22 & 62.88 & 75.24 \\
    \quad w/ ST & 46.86 & 35.13 & 33.56 & 58.62 & 26.69 & \quad w/ ST & 61.34 & 51.96 & 55.69 & 63.12 & 77.41 \\ \midrule
    CSS & 45.69 & 34.59 & 32.73 & 58.69 & 26.67 & CSS & 62.97 & 53.27 & 57.32 & 63.55 & 79.13 \\ \bottomrule
\end{tabular}}
\caption{Domain joint accuracy using 1\% and 5\% labeled training data.}
\label{15avg}
\end{table*}

\begin{table*}[!t]
\centering
\scalebox{0.785}{
\begin{tabular}{@{}l|ccccc||l|ccccc@{}}
    \toprule
    10\% & attraction & hotel & restaurant & taxi & train & 25\% & attraction & hotel & restaurant & taxi & train \\ \midrule
    BASE & 66.82 & 54.50 & 59.97 & 69.33 & 77.03 & BASE & 69.49 & 57.26 & 66.38 & 78.00 & 79.21 \\ 
    \quad w/ SSL & 66.80 & 55.64 & 61.04 & 69.27 & 77.30 & \quad w/ SSL & 71.13 & 57.91 & 66.92 & 78.60 & 78.47 \\
    \quad w/ ST & 68.56 & 56.57 & 62.46 & 69.85 & 80.12 & \quad w/ ST & 71.57 & 59.67 & 66.88 & 78.13 & 81.38 \\ \midrule
    CSS & 68.14 & 57.43 & 64.25 & 70.52 & 79.97 & CSS & 71.83 & 58.01 & 68.41 & 78.43 & 80.89\\ \bottomrule
\end{tabular}}
\caption{Domain joint accuracy using 10\% and 25\% labeled training data.}
\label{1025avg}
\end{table*}

\begin{table*}[!t]
\centering
\scalebox{0.75}{
\begin{tabular}{@{}l|ccccc@{}}
    \toprule
    1\% & attraction & hotel & restaurant & taxi & train \\ \midrule
    BASE & (42.76, 44.40, 42.98) & (35.92, 30.51, 31.04) & (31.38, 33.88, 34.06) & (58.19, 58.77, 58.45) & (24.65, 26.26, 26.11)\\
    \quad w/ SSL & (40.66, 48.05, 41.34) & (32.98, 32.39, 32.42)& (27.81, 32.99, 33.22)& (58.19, 58.13, 59.16) & (23.62, 25.34, 28.14)\\
    \quad w/ ST & (45.79, 49.56, 45.24) & (38.29, 32.67, 34.42) & (30.52, 34.00, 36.17) & (58.77, 58.65, 58.45) & (25.13, 27.85, 27.08)\\ \midrule
    CSS & (43.05, 49.63, 44.40) & (34.95, 35.32, 33.51) & (29.68, 34.80, 33.70) & (58.58, 58.52, 58.97) & (23.46, 26.16, 30.39)\\ \bottomrule
\end{tabular}}
\caption{Domain joint accuracy using 1\% labeled training data on three seeds.}
\label{1shot}
\end{table*}

\begin{table*}[!t]
\centering
\scalebox{0.75}{
\begin{tabular}{@{}l|ccccc@{}}
    \toprule
    5\% & attraction & hotel & restaurant & taxi & train \\ \midrule
    BASE & (59.89, 60.86, 60.02) & (49.11, 49.11, 48.17) & (51.06, 54.51, 53.94) & (62.71, 62.52, 64.13) & (71.71, 71.27, 74.82) \\
    \quad w/ SSL & (60.31, 60.79, 61.83) & (50.64, 51.27, 50.30) & (51.15, 54.24, 54.27) & (61.03, 64.00, 63.61) & (72.25, 75.96, 77.52)\\
    \quad w/ ST & (62.99, 60.44, 60.60) & (53.74, 51.86, 50.27) & (53.89, 55.88, 57.31) & (62.39, 62.90, 64.06) & (77.31, 76.54, 78.40)\\ \midrule
    CSS & (62.79, 63.92, 62.21)  & (55.39, 52.55, 51.86) & (55.73, 58.86, 57.37) & (62.53, 64.06, 64.06) & (79.08, 80.51, 77.81)\\ \bottomrule
\end{tabular}}
\caption{Domain joint accuracy using 5\% labeled training data on three seeds.}
\label{5shot}
\end{table*}

\begin{table*}[!t]
\centering
\scalebox{0.75}{
\begin{tabular}{@{}l|ccccc@{}}
    \toprule
    10\% & attraction & hotel & restaurant & taxi & train \\ \midrule
    BASE & (64.80, 67.15, 68.51)  & (55.49, 55.27, 52.74) & (59.42, 58.65, 61.83) & (70.52, 68.45, 69.03) & (76.89, 76.60, 77.60) \\
    \quad w/ SSL & (64.18, 67.67, 68.54) & (57.17, 56.24, 53.52) & (60.08, 61.18, 61.86) & (68.77, 69.29, 69.74) & (77.20, 76.86, 77.84)\\
    \quad w/ ST & (68.64, 67.34, 69.70) & (58.67, 55.33, 55.70) & (63.05, 62.55, 61.77) & (69.55, 69.35, 70.65) & (79.43, 79.77, 81.15)\\ \midrule
    CSS & (66.57, 68.86, 68.99) & (58.30, 56.86, 57.14) & (62.99, 66.15, 63.62) & (69.68, 70.52, 71.35) & (81.41, 79.43, 79.08)\\ \bottomrule
\end{tabular}}
\caption{Domain joint accuracy using 10\% labeled training data on three seeds.}
\label{10shot}
\end{table*}

\begin{table*}[!t]
\centering
\scalebox{0.75}{
\begin{tabular}{@{}l|ccccc@{}}
    \toprule
    25\% & attraction & hotel & restaurant & taxi & train \\ \midrule
    BASE & (67.86, 69.44, 71.18) & (57.24, 57.49, 57.05) & (68.59, 64.78, 65.76) & (79.42, 76.84, 77.74) & (80.57, 78.10, 78.95)\\
    \quad w/ SSL & (69.54, 72.12, 71.73) & (58.39, 57.77, 57.58) & (66.63, 67.94, 66.18) & (78.58, 78.65, 78.58) & (79.19, 78.66, 77.55)\\
    \quad w/ ST & (71.44, 70.28, 72.99) & (58.77, 60.14, 60.11) & (66.89, 67.85, 65.91) & (77.94, 77.87, 78.58) & (82.03, 81.02, 81.10)\\ \midrule
    CSS & (71.67, 72.93, 70.89) & (56.92, 58.21, 58.89) & (67.52, 69.43, 68.29) & (78.39, 78.52, 78.39) & (81.89, 81.15, 79.64) \\ \bottomrule
\end{tabular}}
\caption{Domain joint accuracy using 25\% labeled training data on three seeds.}
\label{25shot}
\end{table*}

\end{document}